\documentclass[conference]{IEEEtran}
\usepackage{amsmath,amsfonts}
\usepackage{hyperref}
\usepackage{array}
\usepackage[caption=false,font=normalsize,labelfont=sf,textfont=sf]{subfig}
\usepackage{textcomp}
\usepackage{stfloats}
\usepackage{url}
\usepackage{color}
\usepackage{verbatim}
\usepackage{graphicx}
\usepackage{lipsum}
\usepackage{makecell, multirow, booktabs}
\usepackage{caption}
\usepackage[space]{cite}
\usepackage[OT1]{fontenc}

\makeatletter
\newcommand{\removelatexerror}{\let\@latex@error\@gobble}
\makeatother

\usepackage[ruled,norelsize, linesnumbered]{algorithm2e}

\IEEEoverridecommandlockouts

\begin{document}
\bstctlcite{IEEEexample:BSTcontrol}

\title{\LARGE \bf
Multi-modal Integrated Prediction and Decision-making with Adaptive Interaction Modality Explorations
}

\author{Tong Li$^{1*}$, Lu Zhang$^{1*}$, Sikang Liu${^2}$, Shaojie Shen${^1}$
\thanks{$^{*}$Contributed equally to this work.}
\thanks{$^{1}$T. Li, L. Zhang and S. Shen are with the Department of Electronic and Computer Engineering, Hong Kong University of Science and Technology, Hong Kong (e-mail: tlibm@ust.hk; lzhangbz@ust.hk; eeshaojie@ust.hk).}
\thanks{$^{2}$S. Liu is with the DJI Technology Company, Ltd., Shenzhen 518057, China (e-mail: sikang.liu@dji.com).}}

\maketitle

\begin{abstract}
Navigating dense and dynamic environments poses a significant challenge for autonomous driving systems, owing to the intricate nature of multimodal interaction, wherein the actions of various traffic participants and the autonomous vehicle are complex and implicitly coupled. In this paper, we propose a novel framework, \underline{M}ulti-modal \underline{I}ntegrated predictio\underline{N} and \underline{D}ecision-making (MIND), which addresses the challenges by efficiently generating joint predictions and decisions covering multiple distinctive interaction modalities. Specifically, MIND leverages learning-based scenario predictions to obtain integrated predictions and decisions with social-consistent interaction modality and utilizes a modality-aware dynamic branching mechanism to generate scenario trees that efficiently capture the evolutions of distinctive interaction modalities with low variation of interaction uncertainty along the planning horizon. The scenario trees are seamlessly utilized by the contingency planning under interaction uncertainty to obtain clear and considerate maneuvers accounting for multi-modal evolutions. Comprehensive experimental results in the closed-loop simulation based on the real-world driving dataset showcase superior performance to other strong baselines under various driving contexts. Code is available at: \url{https://github.com/HKUST-Aerial-Robotics/MIND}.
\end{abstract}

\section{Introduction}

While autonomous driving technology has made remarkable strides recently, navigating through dense and dynamic traffic remains a formidable challenge. Generating safe and smooth maneuvers in such situations requires accurate modeling of interaction among agents and reasoning about how the scenario evolves in the future, which is non-trivial since the intentions of agents are inherently multimodal and mostly coupled with each other, even with perfect perception results~\cite{ding2021epsilon, cui2019multimodal, hubmann2018automated}.

\indent Extensive research has been conducted to address the challenge by introducing learning-based integrated prediction and planning systems. Some existing approaches adopt explicit hierarchical modeling and address these two tasks separately, with one serving as the conditional input for the other. For instance, following the ``predict-then-plan" pipeline, \cite{zeng2020dsdnet, cui2021lookout, casas2021mp3} generate multimodal motion prediction for all agents in the scene, then leverage them as the input of the following trajectory planning task. On the contrary, \cite{rhinehart2019precog, song2020pip, salzmann2020trajectron++} sample goals or trajectories of the ego vehicle in advance and then use them as additional conditions for the motion prediction network, aiming to model the influence of ego plan on other agents. However, these hierarchical approaches fail to model the implicit bidirectional interaction and would potentially lead to unrealistic predictions and decisions, such as over-conservative or over-optimistic behaviors. Recently, joint multi-agent motion forecasting models~\cite{casas2020implicit, girgis2021latent, ngiam2021scene, kang2023ffinet} have been widely studied, which focus on predicting multiple possible future scenarios that are physically and socially consistent given the driving context. These approaches employ deep neural networks to implicitly capture the inherent dependencies and interactions among agents. Typically, when the ego vehicle is integrated into the model, the network is capable of predicting its future trajectories as well. Although these predictions could inform the ego vehicle's decisions to some extent, directly using them as such results in undesired performance. One reason is that, despite modern networks' ability to model interactions between agents and static scenes effectively, long-term and scene-consistent prediction remains difficult~\cite{casas2020implicit, kang2023ffinet}. The uncertainty of predictions always escalates after just a few seconds due to the inherent multimodality, which brings unreliable decisions. On the other hand, plain joint motion forecasting models struggle to produce desired trajectories for the ego vehicle without extra guidance, highlighting a limitation in their application~\cite{hagedorn2023rethinking, chen2023tree}.

\begin{figure}[!t]
\centering
\includegraphics[width=0.48\textwidth]{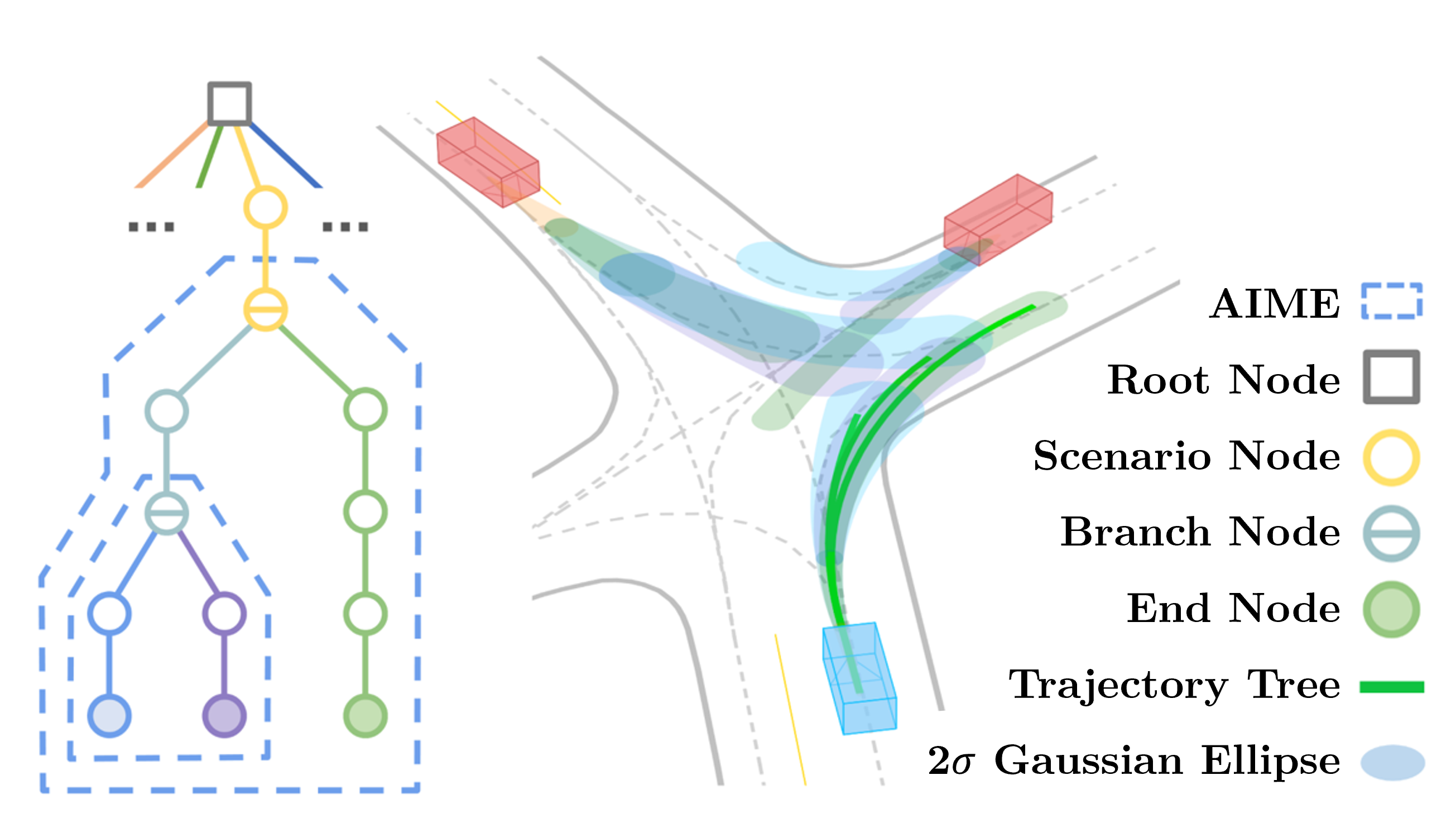}
\caption{A branch of the generated scenario tree and its corresponding topological structure. In MIND, we employ a learning-based scene-consistent driver model coupled with the adaptive interaction modality exploration (AIME) mechanism to efficiently construct the scenario tree. For each branch originating from the root node, we utilize contingency planning to generate a trajectory tree, accommodating multi-modal future evolutions.}
\label{fig:cover}
\vspace{-0.6cm}
\end{figure}

For sequential decision-making, tree search has been widely applied, which considers the dynamics of the world and investigates its evolution to the future. Traditional decision-making approaches frame problems as partially observable Markov decision processes (POMDPs) and employ tree search techniques to derive sub-optimal solutions~\cite{luo2018porca, hubmann2018automated}. Given the challenges in scaling and the sometimes intractable nature of modeling behaviors and interactions of driving agents in complex environments through human heuristics and handcrafted rules, contemporary approaches incorporate neural networks to model transition and observation functions~\cite{chen2023tree, huang2023dtpp, chekroun2023mbappe}. This adaptation facilitates the generation of human-like interactive driving maneuvers in various situations. However, existing methods often rely on a fixed tree structure or explicitly decouple the prediction and planning processes, resulting in limited flexibility.

\indent To address the above limitations, we introduce the \underline{M}ulti-modal \underline{I}ntegrated predictio\underline{N} and \underline{D}ecision-making (MIND), a novel method that systematically combines a learning-based integrated prediction and planning model, a dynamic branching mechanism, and contingency planning on multi-modal future evolutions, enabling the generation of reasonable behaviors in complex interaction scenarios. In MIND, we adopt a lightweight and efficient joint multi-agent motion prediction network designed to produce scene-consistent future distributions for both the ego vehicle and all surrounding agents. In the planning phase, this network acts as the ``world dynamics", enabling the construction of scenario trees by its recursive invoking. To ensure comprehensive coverage of pivotal scenarios while avoiding redundant search efforts, we introduce a dynamic branching mechanism named Adaptive Interaction Modality Exploration (AIME), which utilizes the uncertainty in the predicted states of agents to guide the branching process. To identify the optimal decision, we evaluate each branch originating from the root node and find the most advantageous one. In line with our previous work~\cite{li2023marc}, we determine the policy by evaluating both scenario and trajectory trees produced by contingency planning. This approach allows for adherence to various constraints and cost functions, thereby improving the ability to handle scene uncertainty. A typical illustration can be found in Fig.~\ref{fig:cover}. We validate the effectiveness of MIND by conducting comprehensive experiments via open-loop and closed-loop simulations. The results demonstrate its superior performance in diverse driving scenes compared to other baselines, underscoring its potential to facilitate autonomous driving in complex environments. We summarize the contributions of this paper as follows:
\begin{itemize}
    \item We design a scene prediction network and integrate it with tree search techniques featuring a dynamic branching mechanism, resulting in a scenario tree with enhanced coverage for exploring the world's evolutions.
    \item For the multiple potential futures within the scenario tree, we utilize contingency planning to naturally generate optimal trajectory trees against each branch originating from the root, thereby determining the best decision.
    \item We evaluate MIND through various experiments, with results outperforming other baselines across diverse driving scenarios, showing its efficacy in complex situations.
\end{itemize}

\section{Related Work}\label{sec:related_work}

\subsection{Joint Multi-agent Motion Prediction}

Previous studies mostly focus on predicting a single target agent given its surrounding context~\cite{liang2020learning, gao2020vectornet, varadarajan2022multipath++}. However, considering that the behaviors of road users are interdependent, achieving high prediction accuracy for a single agent is insufficient, while ensuring physical and social consistency among all participants is equally paramount. For multi-agent joint prediction, \cite{zeng2020dsdnet, luo2023jfp} initially generate marginal predictions for each agent, which are then integrated using a deep structured model to deduce the joint distribution of behaviors. Factorization-based approaches \cite{sun2022m2i, rowe2023fjmp} tackle the joint prediction by explicitly establishing a partial order of target agents, and then modeling the problem as a conditional prediction task. Implicit methods~\cite{casas2020implicit, girgis2021latent, ngiam2021scene, kang2023ffinet}, by directly forecasting the joint possible future with minimal assumptions and inductive biases, offer enhanced generalizability and improved computational efficiency. We follow the implicit methods and adopt an efficient joint motion predictor based on our previous work~\cite{zhang2024simpl}.

\subsection{Integrated Prediction and Planning}
Partially observable Markov decision process (POMDP) offers a mathematically rigorous approach to modeling uncertainties and multi-agent interactions but is challenged by its high computational complexity. Even with efficient POMDP solvers, existing methods~\cite{luo2018porca, hubmann2018automated} struggle to meet the real-time requirements of decision-making tasks in autonomous driving. Some approaches achieve satisfactory results in real systems by simplifying the original problem using domain knowledge~\cite{cunningham2015mpdm, zhang2020efficient}. However, these systems grounded in human experience often exhibit limited flexibility in complex environments and pose challenges for scalability. Alongside, there's a pivot towards learning-based integrated systems for overcoming prediction and planning challenges. These systems vary from hierarchical models, where prediction sequentially informs planning~\cite{zeng2020dsdnet, cui2021lookout, casas2021mp3}, to approaches that integrate the ego vehicle's intended actions into motion prediction, considering their effects on other agents~\cite{rhinehart2019precog, song2020pip}. Despite advancements, such models often miss capturing bidirectional agent interactions, occasionally resulting in impractical behaviors. Additionally, recent methods~\cite{chen2023tree, huang2023dtpp, chekroun2023mbappe} incorporating neural networks for modeling the ``world dynamics" show promise for creating realistic, interactive maneuvers. In this paper, we follow and enhance this pipeline by introducing an adaptive branching mechanism, leading to higher flexibility and efficiency.

\subsection{Motion Planning with Contingency}
Contingency planning is introduced to produce deterministic actions that consider the motion uncertainty of other agents~\cite{hardy2013contingency}. To address potential changes in the intentions of others, \cite{chen2023interactive, wang2023interaction} implement a scenario tree with a predefined topological structure. This is succeeded by optimization over the scenario tree using model predictive control to derive a trajectory tree that captures reactive behaviors in future steps. Combined with multipolicy decision-making~\cite{cunningham2015mpdm}, \cite{li2023marc} introduces risk-aware contingency planning on policy-conditioned scenario trees with dynamic branching points for each policy. In this paper, we extend this idea by optimizing a multi-way scenario tree dynamically constructed by a neural network to ascertain the optimal decision for the ego vehicle.


\section{Framework Overview}\label{sec:sys_overview}
\begin{figure}[!t]
    \centering
    \includegraphics[width=0.475\textwidth]{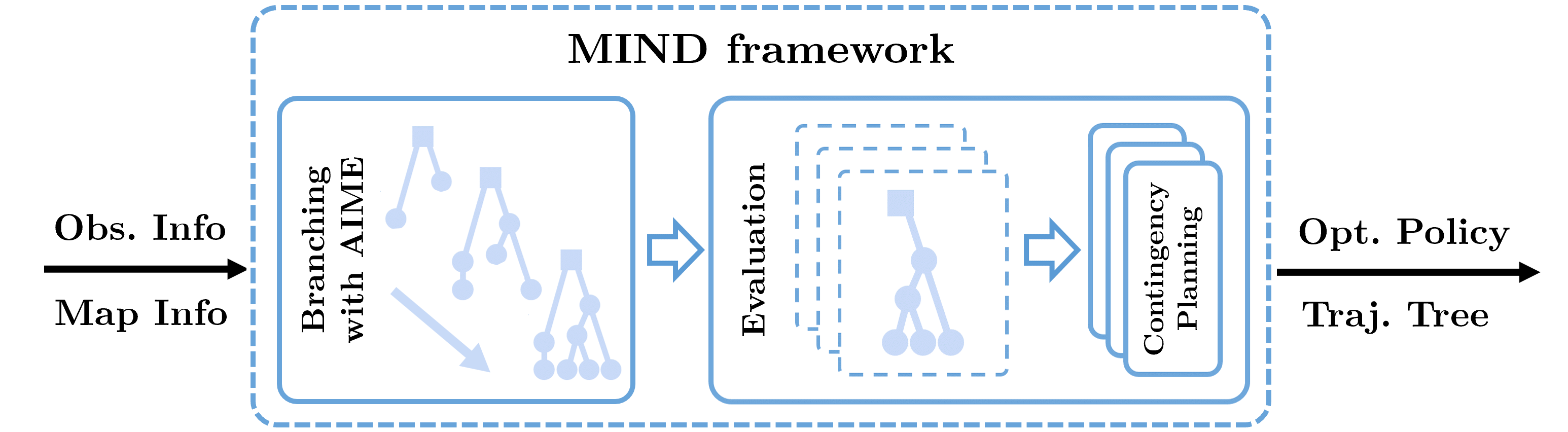}
    \caption{Illustration of the components and the workflow in the MIND framework.}
    \label{fig:framework}
    \vspace{-0.6cm}
\end{figure}

\autoref{fig:framework} depicts the proposed MIND framework, which consists of two key procedures: \textit{dynamically building the scenario tree with AIME} and \textit{policy evaluation}. Utilizing the observations and environmental data, MIND efficiently creates a scenario tree through AIME-guided branching. The tree explores the future interaction modalities, incorporating both predictions and decision-making processes simultaneously. The selection of the optimal policy is determined by assessing the scenario evolutions within each branch that stems from the root node, along with the trajectory tree obtained from contingency planning for handling multi-modal future interactions. Details are provided in Sec.~\ref{sec:aime} to Sec.~\ref{sec:imple_detail}.

\section{Adaptive Interaction Modality Explorations}\label{sec:aime}
\subsection{Integrated Prediction and Decision}
As elaborated in Sec.~\ref{sec:related_work}, learning-based methods can generate the joint predictions of agents and the ego vehicle, namely scenario predictions, based on the driving context. In MIND, we leverage a transformer-based network to anticipate the scenario in the form of \textit{Gaussian mixture models} (GMMs). We first denote the map information as~$\mathbf{M}$, the historical observations~$e$ as~$\mathbf{X}$ which contain the observed trajectories of~$N_a$ moving agents and ego vehicle over the past~$H$ time steps. The network generates joint distributions~$\mathbf{Y}$ consisting of agents' predictions and ego decisions in the horizon $T$:
\begin{gather}
P(\mathbf{Y} \mid \mathbf{X}, \mathbf{M})= P\left(\mathbf{Y} \mid \mathbf{Z}, \mathbf{X}, \mathbf{M}\right)P\left(\mathbf{Z} \mid \mathbf{X}, \mathbf{M}\right),
\end{gather}
where~$\mathbf{Z}$ are latent variables that capture unobserved features (e.g., agent intentions, driving styles, and interactions). To ensure clarity, we define the scenario node~$\mathbf{Y}_t^{k}$ of the~$k$-th predicted scenario at time step~$t$ as follows:
\begin{gather}
P\left(\mathbf{Y}_{t}^{k} \mid \mathbf{X}, \mathbf{M}\right)= \alpha_k \mathcal{N}\left(\mu^{k, i}_{t}, \Sigma^{k, i}_{t}\right), i \in \{e, 1, ..., N_a\},
\end{gather}
with~$\alpha_k$, ~$\mu^{k, i}_t$ and~$\Sigma^{k, i}_t$ representing the probability score, mean and covariance of the positional Gaussian~$\mathcal{N}$ of a certain agent or ego vehicle, respectively. The joint distribution of agents' positions at time step~$t$ can be expressed as a GMM:
\begin{gather}
P\left(\mathbf{Y}_{t} \mid \mathbf{X}, \mathbf{M}\right)= \textstyle \sum_{k=1}^K P\left(\mathbf{Y}_{t}^{k} \mid \mathbf{X}, \mathbf{M}\right),
\end{gather}
The~$k$-th predicted scenario~$\mathbf{Y}^{k}$ and the overall scenario predictions~$\mathbf{Y}$ are defined as below:
\begin{gather}
\mathbf{Y}^{k} = \{\mathbf{Y}_{t}^{k}\}, \ t \in \{1, ..., T\},\\
\mathbf{Y} = \{\mathbf{Y}^{k}\}, \ k \in \{1, ..., K\}.
\end{gather}
To be more specific, the network estimates the Gaussian over actions of the single integrator, which are then converted into spatial predictions through linear Gaussian dynamics. We simplify the description here for clarity. For further detail on this conversion process, we kindly refer interested readers to~\cite{salzmann2020trajectron++}. Under the linear dynamic assumption, the positional distribution of any agents at a given future time step can be obtained by propagating from the previous positional Gaussian and the predicted action Gaussian, facilitating the recursive branching for the expansions of the scenario tree. 

\indent To improve the long-term prediction accuracy, the network can also incorporate high-level planning commands which are defined as the intended route in MIND. These commands, which can be aligned with ground truth during training and generated on board, are introduced into the scenario decoder rather than the fusion network to prevent bias in agent interactions. The network design and implementations are further detailed in Sec.~\ref{sec:imple_detail}. Note that while the high-level commands serve as options for enhanced conditioned predictions, the long-term generations of desired ego decisions are primarily guided by the subsequent pruning and merging procedure. The effectiveness of the network in both unconditioned and conditioned scenario prediction tasks is evaluated in Sec.~\ref{sec:exp_result}.

\subsection{Branching Decision based on Uncertainty Variation}
Accurately anticipating the scenarios with single-shot predictions is difficult due to the agent intentions' multimodality and coupling over time and situations~\cite{casas2020implicit, cui2019multimodal}. With the ``world dynamics", the implicit transition and observation functions of both agents and the ego vehicle learnt by the network, it's intuitive to explore multiple possible evolutions at different time steps, namely branching, to obtain distinctive joint distributions under different interactions. However, brute-force branching with a fixed time interval leads to computational inefficiency and exponential complexity. Recognizing that interaction changes affect agents' future actions, manifesting as increased covariance in GMMs, we conduct an evaluation on~$\mathbf{Y}^{k}$ to dynamically determine a branching time step~$t_b^k$:
\begin{gather}
t_{b}^{k} =\underset{t}{\operatorname{argmax}} \enspace \mathcal{U}(\mathbf{Y}_{t}^{k}) < \beta, \quad t \in \mathcal{Z}^{+},
\vspace{-0.6cm}
\end{gather}
where $\mathcal{U}$ is the measuring function which evaluates the change rate of variation and $\beta$ is a customized tolerance of uncertainty achieving the trade-offs between evolution diversity and computational efficiency. If the determined branching point $t_{b}$ falls within the planning horizon~$T$, a branching process is executed. This branching process updates a pseudo-observation~$\bar{\mathbf{X}}$ with the means of current predicted GMM components and generates the consequent possible scenarios~$P\left(\mathbf{Y} \mid \bar{\mathbf{X}}, \mathbf{M}\right)$ leveraging the prediction network and the attributes of GMMs under linear dynamics. If no branching point is found within $T$, the predictions are truncated to~$T$ and marked as the end scenarios. With this general strategy, the branching decision can effectively adapt the scenario tree to diverse situations.

\begin{figure}[t]
\centering
\includegraphics[width=0.485\textwidth]{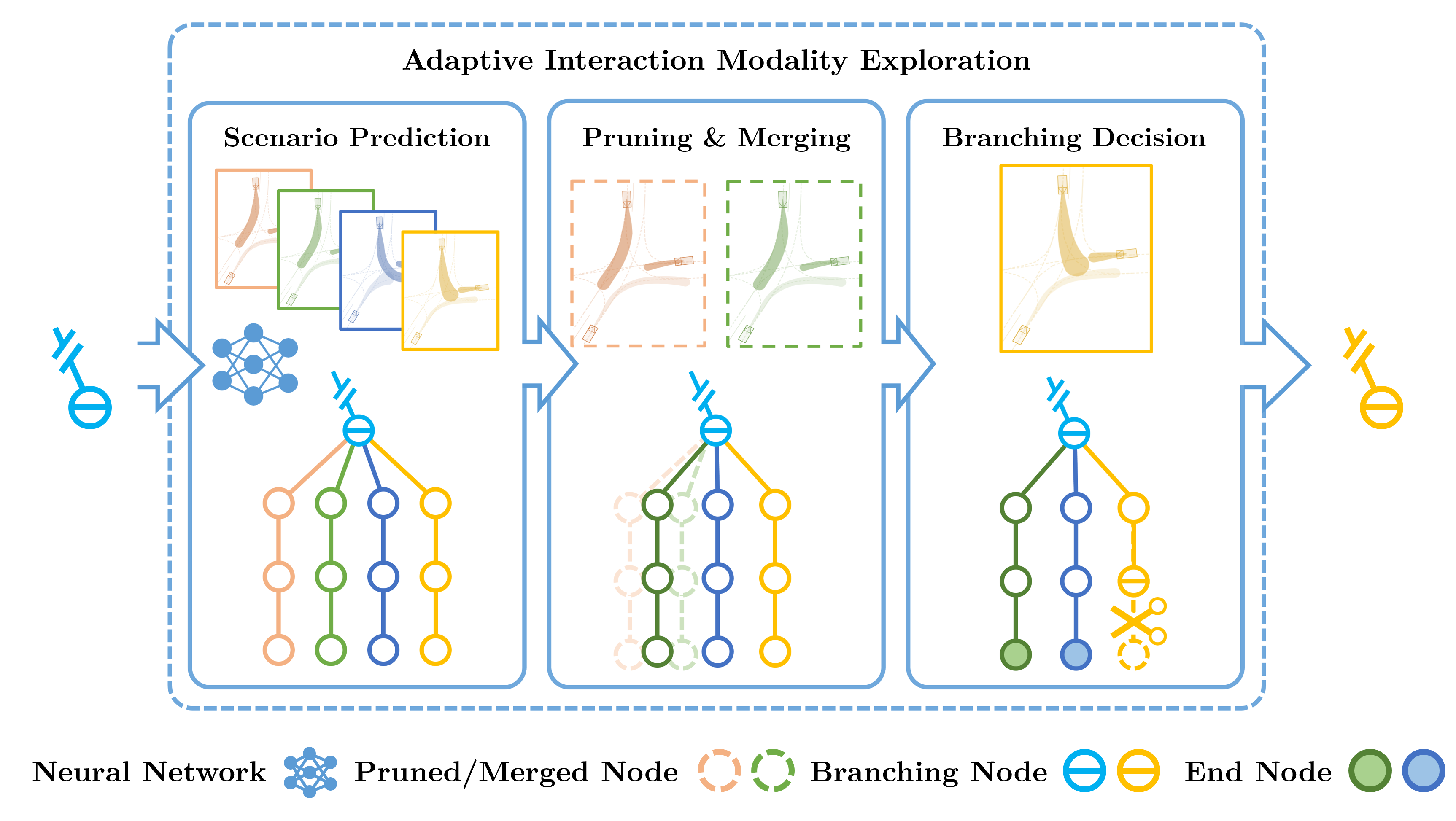}
\caption{Illustration of one AIME-guided branching. The nodes of the scenario tree contain the states of the ego vehicle and agents. Firstly, the scenario tree is extended on the branching node leveraging the scenario prediction network. Then, the extended scenario tree is simplified by the pruning and merging process according to the interaction modality analysis. Finally, end nodes and branching nodes are determined during the adaptive branching, which triggers the next AIME process if branching nodes exist.}
\label{fig:ime}
\vspace{-0.6cm}
\end{figure}
\subsection{Pruning and Merging based on Interaction Modality}
The network's multimodal nature might inadvertently lead to the generation of undesired maneuvers, potentially compromising long-term decisions. Additionally, preserving all similar expanded scenarios without differentiation leads to exponentially increasing computational complexity when extracting critical actions from scenario trees. Therefore, the predicted scenario should be further evaluated and processed, guiding the development of the scenario tree towards the desired evolutions. We introduce a pruning process to discard scenarios with deviated ego decisions and low probabilities. On the other hand, we observe that the agents in similar scenarios exhibit alike interactions and end with adjacent final positions, which can be well categorized by the recently proposed free-end homotopy~\cite{chen2023interactive}. For an ego-agent pair, the homotopy class~$h^{e \rightarrow i}$ is defined as follows:
\begin{gather}
\Delta d^{e \rightarrow i}=\textstyle \sum \left(d^{e \rightarrow i}_{t}-d^{e \rightarrow i}_{t-1}\right)_{\sim}, h^{e \rightarrow i}=\left\lfloor\frac{\Delta d^{e \rightarrow i}}{\delta}+\frac{1}{2}\right\rfloor,
\end{gather}
where~$d^{e \rightarrow i}_{t}$ is the angle between the mean positions of an ego-agent pair at time step~$t$, $\left(\cdot\right)_{\sim}$ normalizes the angle difference to~$(-\pi, \pi]$, $\delta$ is the quantization factor of homotopy class, $\lfloor \cdot \rfloor$ is the floor function that rounds the resulting value down to the nearest integer. We define the interaction modality~$I$ of scenario~$\mathbf{Y}$ based on the homotopy classes of ego-agent pairs:
\begin{gather}
I(\mathbf{Y}) :=\left(h^{e \rightarrow 1}, ..., h^{e \rightarrow N_a} \right).
\end{gather}
Given the scenarios of the same interaction modality, the merging process picks the one with the highest probability as the representative and discards the others while summing up their probabilities. The probabilities of scenarios are normalized across the scenario tree to ensure consistency.

\indent Multiple possible evolutions with distinctive joint distributions are gradually revealed by repeatedly executing branching, pruning, and merging, forming a scenario tree that adaptively explores the interaction modality space. Consequently, we term this comprehensive procedure Adaptive Interaction Modality Explorations (AIME). Scenario trees guided by AIME have more distinctive agent behaviors with lower uncertainties in each time step, benefiting the following contingency planning while preserving compact structures for computation efficiency. In practice, we assign a maximum branch depth~$d_{max}$ to AIME and abandon the rare interaction modalities, which have high uncertainties and need excessive branching to reveal.  An AIME iteration is illustrated in~\autoref{fig:ime} and the complete methodology is detailed in~\autoref{alg:1}.

\begin{figure}[!t]
\removelatexerror
\SetKwInOut{Input}{Inputs}
\SetKwInOut{Output}{Outputs}
\SetKwComment{Comment}{//}{}
\SetKwComment{SComment}{/* }{}
\begin{algorithm}[H]
\LinesNumbered
\caption{Branching with AIME}\label{alg:1}
\Input{$\mathbf{M}$,~$\mathbf{X}$,~$T$,~$\beta$,~$\delta$,~$d_{max}$ }
\Output{Scenario Tree~$\Psi$}
$N_{0} \gets \{\mathbf{X}, 0\}, \mathbf{E} \gets \emptyset, \mathbf{B} \gets \{N_{0}\}$ \Comment*[l]{ Init Node}
\While{$\mathbf{B} \neq \emptyset$}{
    $\bar{\mathbf{B}} \gets\emptyset$ \;
    \For{$N \in \mathbf{B}$}{
        \SComment{AIME Iteration */}
        $\mathbf{X}, d \gets N$ \;
        \If{$d \leq d_{max}$}{
            $\mathbf{Y} \gets $ ScenarioPrediction($\mathbf{X},~\mathbf{M}$) \;
            $\bar{\mathbf{Y}} \gets $ Pruning\&Merging($\mathbf{Y}, \delta$) \;
            \For{$\mathbf{Y}^k \in \bar{\mathbf{Y}}$}{
                \SComment{Branching Decision */}
                $t_{b}^{k} \gets$ GetBranchTime($\mathbf{Y}^k, \beta$) \;
                \eIf{$t_{b}^{k} < T$}{
                $\bar{\mathbf{X}} \gets $ UpdateObser($\mathbf{X}, \mathbf{Y}^k$) \;
               }{
                $\bar{\mathbf{X}} \gets $ TruncatePred($\mathbf{X}, \mathbf{Y}^k$) \;
               }
                 \SComment{Create New Node */}
                $\bar{N} \gets \{\bar{\mathbf{X}}, d+1\}$ \& AddNode($\bar{N}$, $N$)  \;
                \eIf{$t_{b}^{k} < T$}{
                $\bar{\mathbf{B}} \gets \bar{\mathbf{B}} + \{\bar{N}\}$ \;
               }{
                $\mathbf{E} \gets \mathbf{E} + \{\bar{N}\}$ \;
               }
            }
        }
    }
    $\mathbf{B} \gets \bar{\mathbf{B}}$ \;
}
$\Psi \gets $ GetScenarioTree($\mathbf{E}$) \;
\end{algorithm}
\vspace{-0.6cm}
\end{figure}

\section{Evaluation via Contingency Planning}\label{sec:contin_plan}
As the scenario tree unfolds across various interaction modalities over time, policies that handle diverse future evolutions naturally arise. Specifically, we define the GMM ego decision sequence, spanning from the root node to the end nodes in the sub-tree, as a policy generated by the AIME-guided scenario tree, illustrated by the expanded sub-tree in~\autoref{fig:cover}. For a given policy, it is necessary to determine deterministic actions that effectively address multimodal agent predictions for further evaluation and execution. Following our previous work~\cite{li2023marc}, we utilize contingency planning, a compact yet efficient solution for handling multiple evolutions. We extend this technique to incorporate integrated decisions and predictions in scenario trees in the form of GMMs. We denote the number of predicted scenarios in the sub-tree by~$N_s$. Given the~$j$-th scenario, we denote the index of its preceding scenario by~$\bar{j}$, its branch time by~$t_{b}^{j}$, the full time step set of it by~$T_j=\{t_{b}^{\bar{j}}+1, ..., t_{b}^{j}\}$ and a finite set that exclude the first time step~$T_j^{-} = T_j \backslash \{t_{b}^{\bar{j}}+1\}$. For~$j=1$, we have~$t_{b}^{\bar{j}}=0$, which refers to the time step of the root node. With a slight abuse of notation, we define the state and control of the trajectory tree in the~$j$-th scenario at time step~$t$ by $x_{t}^{j}$ and $u_{t}^{j}$, the set of states by~$X$ and the set of control actions by~$U$. The trajectory tree $\tau$ is obtained as follows:
\begin{align}&\tau := \min_{U} \textstyle \sum_{j=1}^{N_s} \textstyle \sum_{t\in T_j} (l_{t}^{j}(x_{t}^{j}, u_{t}^{j}) + \gamma n_{t}^{j}(x_{t}^{j})) \quad \quad \\
\text{s.t.} \ &x_{1}^{1} = f(\hat{x_{0}}, u_{1}^{1}), \ x_{t_{b}^{\bar{j}}+1}^{j} = f(x_{\bar{j}}^{\bar{j}}, u_{\bar{j}+1}^{j}), \ \forall j \in \{2, ..., N_s\}, \nonumber\\ 
&x_{t}^{j} = f(x_{t-1}^{j}, u_{t}^{j}), \ t \in T_j^{-}, \forall j \in \{1, ..., N_s\}, \nonumber\\ 
&h_{t}^{j}(x_{t}^{j}, u_{t}^{j}) \leq \vec{0}, \quad t \in T_j, \forall j \in \{1, ..., N_s\}, \nonumber\\
&\mathcal{P}\{g_{t}^{j}(x_{t}^{j}, u_{t}^{j}) \leq 0\} \geq 1 - p, t \in T_j, \forall j \in \{1, ..., N_s\}, \nonumber
\end{align}
where~$n_{t}^{j}$ is the \textit{negative log-likelihood} (NLL) loss with respect to the Gaussian distribution of the ego decision weighted by non-negative factor~$\gamma$. $f(\cdot)$ is the state-transition function, $h_{t}^{j}(\cdot)$ is the deterministic multi-dimensional constraint function, $l_{t}^{j}(\cdot)$ is the customized loss, the first three constraints ensure the trajectory tree starts from the state of root node~$\hat{x_{0}}$ and the continuity of the trajectory tree within the scenarios and between scenarios and their predecessor, and $g_{t}^{j}(\cdot)$ is the safety constraint function defined on the Gaussian distributions of agent predictions. $\mathcal{P}$ returns the probability of the input function, and $p$ is a tolerance of constraint violation probability. The problem above can be integrated with risk measurement such as \textit{conditional value-at-risk} to form a \textit{risk-aware contingency planning} problem~\cite{li2023marc}, aiming to develop actions averse to potential dangers. \\
\indent Solving this extended problem given different sub-trees obtains the trajectory trees with maneuvers accounting for multi-modal under associated policies. The optimal policy and trajectory tree are then chosen based on the reward evaluation:
\begin{gather}
\mathcal{Q}(\tau):= \textstyle \sum_{j=1}^{N_s} \textstyle \sum_{t\in T_j} R(x_{t}^{j}, u_{t}^{j}), \\
\tau^* = \underset{\tau}{\operatorname{argmax}} \enspace \mathcal{Q}(\tau_i), i \in \{1, ..., N_{\tau}\},
\end{gather}
in which~$\mathcal{Q}(\cdot)$ computes the summation of reward function~$R$ of every state in the trajectory tree. The customized reward function is detailed in Sec.~\ref{sec:imple_detail}.

\section{Implementation Details}\label{sec:imple_detail}
\subsection{Scene-level Prediction Network}
\begin{figure}[t]
\centering
\includegraphics[width=0.485\textwidth]{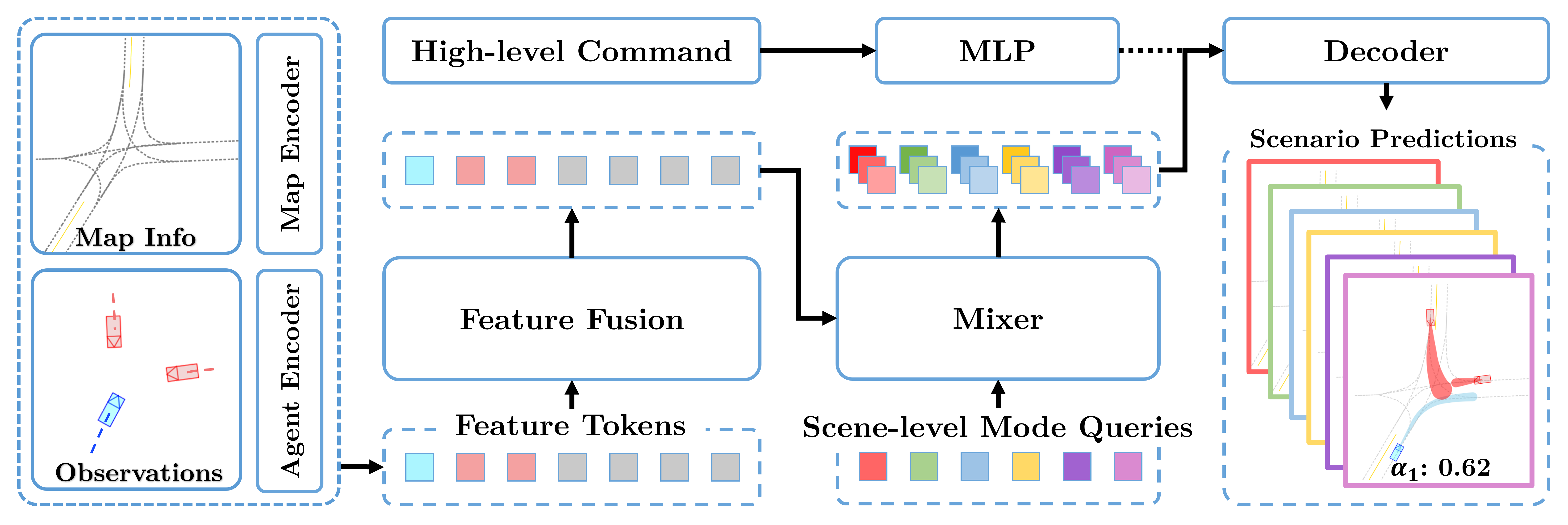}
\caption{Architecture of the scene prediction network. After the feature encoding, we mix the scene-level mode queries with agent features. As illustrated, $A=3$ agents are included in this scene, while $K=6$ mode queries are injected. At last, the scene decoder generates $K$ possible joint future scenes with estimated probabilities.}

\label{fig:network}
\vspace{-0.6cm}
\end{figure}

As shown in~\autoref{fig:network}, the prediction network follows the encoder-decoder architecture that takes the map info and historical observations as inputs and generates multiple future scenes and their probability scores. For the context encoding part, we adopt our previous work~\cite{zhang2024simpl}, which first encodes the observed trajectories of surrounding road users and map elements, and then performs efficient global feature fusion using a Transformer-like network.
To achieve consistent scene prediction, we introduce $K$ scene-level mode queries in the decoding procedure, which represents different interaction modalities or ``consensus" among traffic participants. We mix the mode queries with all agent features and send them to an MLP-based scene decoder to generate $K$ possible future scenes. Specifically, in a driving scenario with $A$ agents, we replicate the fused agent features $K$ times and the mode queries $A$ times, respectively. After that, both tensors have the shape of [$K$, $A$, $D$], where $D$ is the latent size, and we can then aggregate them to get the agent features under each scene. Moreover, we parameterize the predicted trajectories using GMMs in the decoder, reflecting the motion uncertainty.
As mentioned in Sec.~\ref{sec:aime}, the high-level commands, if given, are encoded and directly injected into the scene decoder to obtain the conditioned prediction of ego decisions. This approach avoids involving high-level commands in global feature fusion, namely, the commands of the ego vehicle only affect its own fused features, thereby preventing incorrect message passing and unnecessary dependencies.

We train this network in an end-to-end manner, leveraging a combined regression and classification loss. We utilize the scene-level winner-takes-all strategy~\cite{ngiam2021scene} to avoid mode collapse. As for the classification loss, we employ the max-margin loss~\cite{liang2020learning} to distinguish the winner scene from others.

\subsection{iLQR Design}
We solve the contingency problem utilizing the \textit{iterative linear quadratic regulator} (iLQR)~\cite{mayne1966second}. We adopt the discrete bicycle kinematic model as $f(\cdot)$ design the loss function of iLQR, as shown below:
\begin{gather}
l_t = l_{t}^{\text{safe}} + l_{t}^{\text{tar}} + l_{t}^{\text{kin}} + l_{t}^{\text{comf}} + l_{t}^{\text{dec}} + l_{t}^{\text{col}},
\end{gather}
in which~$l_{t}^{\text{safe}}$, ~$l_{t}^{\text{tar}}$, ~$l_{i}^{\text{kin}}$, and~$l_{i}^{\text{comf}}$ are the safety cost, target cost, kinematic cost, and comfort cost respectively. These cost components align with those defined in MARC~\cite{li2023marc}. To tailor iLQR for the extended contingency planning problem, we incorporate two additional cost elements: $l_{t}^{\text{dec}}$, which calculates the NLL loss on the ego decisions' GMMs, and $l_{t}^{\text{col}}$, which penalizes the potential collision based on GMM predictions. We define $l_{t}^{\text{dec}}$ using the Mahalanobis distance measure function $\mathcal{D}(\cdot)$ on the Gaussian distribution, as the squared of Mahalanobis distance is proportional to the NLL:
\begin{gather}
l_{t}^{\text{dec}} = \mathcal{D}^{2}(\mathcal{N}_{t}^{e}).
\end{gather}
For enforcing safety constraints, limiting collision probabilities with other agents is effectively managed by setting minimum thresholds for the Mahalanobis distance. Consequently, we specify the potential collision penalty $l_{t}^{\text{col}}$ as:
\begin{gather}
l_{t}^{\text{col}} = \textstyle \sum_{j=1}^{N_a} \mathcal{G}\big(\text{max}(\mathcal{D}_{\text{bnd}} - \mathcal{D}(\mathcal{N}_{t}^{j}), 0)\big),
\end{gather}
where $\mathcal{G}(\cdot)$ represents the custom penalty function and $\mathcal{D}_{\text{bnd}}$ denotes the threshold for the Mahalanobis distance such that~$\mathcal{P}\{D \leq \mathcal{D}_{\text{bnd}}\} = 1-p$ for the distribution $\mathcal{N}_{t}^{\mathrm{j}}$ from the GMMs of the $j$-th agent's prediction at time step $t$.

\subsection{Reward Function}
To select trajectory trees that effectively balance efficiency, comfort, and commonness, we propose a multi-dimensional reward function to evaluate the states, controls, and probabilities associated with the trajectory tree~$\tau$ outlined below:
\begin{gather}
R(x_{t}^{j}, u_{t}^{j}) = \lambda_{p}(\lambda_1 F_s + \lambda_2 F_e + \lambda_3 F_c),
\end{gather}
where~$\lambda_{p}$ likelihood-related weight for customized preference on commonness, $\lambda_{1}$, $\lambda_{2}$, and $\lambda_{3}$ denote non-negative weights, $F_s$ assesses the safety by evaluating the Mahalanobis distance to other agents' predictions, $F_e$ evaluates efficiency by comparing the planned velocity against the target velocity, and $F_c$ quantifies comfort based on the planned control.


\section{Experimental Results}\label{sec:exp_result}

\subsection{Experiment Setup}
\subsubsection{Dataset and simulations}
Our experiments are conducted on the Argoverse 2~\cite{wilson2023argoverse} motion forecasting dataset, which offers 10-Hz sequences including 5 seconds of historical data and 6 seconds of future motion predictions, and accompanied by high-definition maps. We conduct multi-agent trajectory prediction evaluation, effectiveness analysis, and closed-loop simulations based on the Argoverse dataset.

\subsubsection{Metrics}
For prediction evaluations, we utilize standard metrics for multi-agent trajectory predictions: average minimum average displacement error (MinADE), average minimum final displacement error (MinFDE), actor miss rate (actorMR), and actor collision rate (actorCR)~\cite{wilson2023argoverse}. The MinADE measures the mean lowest L2 norm to the ground truth, while MinFDE focuses on the endpoint error. The actorMR measures the average deviation ratio of predictions for each scored agent across the evaluation set. The actorCR is the ratio of collisions among agents within the scenario of the lowest MinFDE. For effectiveness analysis of AIME, we evaluate the modality coverage, number of predicted scenarios in the scenario tree, and relative computational cost. For closed-loop simulation evaluations, we utilize typical planning metrics: average speed, maximum absolute acceleration, and root-mean-squared acceleration. The avgSpd measures the overall efficiency, maxAbsAcc captures the uncomfortable maneuvers, and rmsAcc reflects the decision consistency.

\subsubsection{Baseline, Platform and Environment}
For quantitative closed-loop comparisons, we benchmark against two models: a model-based prediction and decision-making module with contingency planning (MD+CP) similar to MARC~\cite{li2023marc} and a learning-based variant in which single-shot results are used for prediction and decision (NN+CP). Both neural networks in NN+CP and MIND are aligned for equitable comparison. The closed-loop experiments are conducted on a self-built multi-agent platform based on the Argoverse 2 dataset. This platform operates synchronously, in which the perception is rendered according to the vehicles' positions and the observations from the data, and the states of the simulated vehicles are updated according to the kinematic model and planned trajectories in each step. Our experiments, including baselines, our proposed system, and the simulation platform, are implemented in Python3. Closed-loop simulations are run on a desktop with an Intel i5-12500KF CPU and an Nvidia RTX 3060 GPU. Network training is conducted with a batch size 128 for 50 epochs on a server with 8 Nvidia RTX 3090 GPUs.

\subsection{Results}
\subsubsection{Quantitative comparison with state-of-the-arts on multi-agent prediction task}
We compare the proposed network with other state-of-the-art methods based on scene-level interaction modeling to validate its performance. Given that our network generates GMM-based outputs that differ from the trajectory outputs typical of multi-agent forecasting tasks, we recalibrate by mapping the GMM predictions to trajectories using the predicted means and probabilities. The quantitative results of the multi-agent motion forecasting benchmark on Argoverse 2 are shown in~\autoref{table_1}. Despite not primarily targeting trajectory predictions, our method outperforms baseline methods, showcasing its efficacy in multi-agent interaction modeling.

\begin{table}[h]
\caption{Results on the test split of the Argoverse 2 multi-world forecasting benchmark. The best result is in \textbf{bold}.} \label{table_1}
\centering
\setlength{\tabcolsep}{1.8mm}
\begin{tabular}{l|cccc} 
\toprule
Methods                      & MinADE$_6$ & MinFDE$_6$ & actorMR$_6$    & actorCR$_6$ \\ 
\midrule
FJMP~\cite{rowe2023fjmp}     & 0.81          & 1.89          & 0.23           & 0.01 \\ 
FFINet~\cite{kang2023ffinet} & 0.77          & 1.77          & 0.24           & 0.02 \\ 
Proposed                     & \textbf{0.70} & \textbf{1.62} & \textbf{0.20}  & \textbf{0.009} \\ 
\bottomrule
\end{tabular}
\vspace{-0.6cm}
\end{table}

\begin{figure}[tbh]
    \centering
    \includegraphics[width=0.48\textwidth]{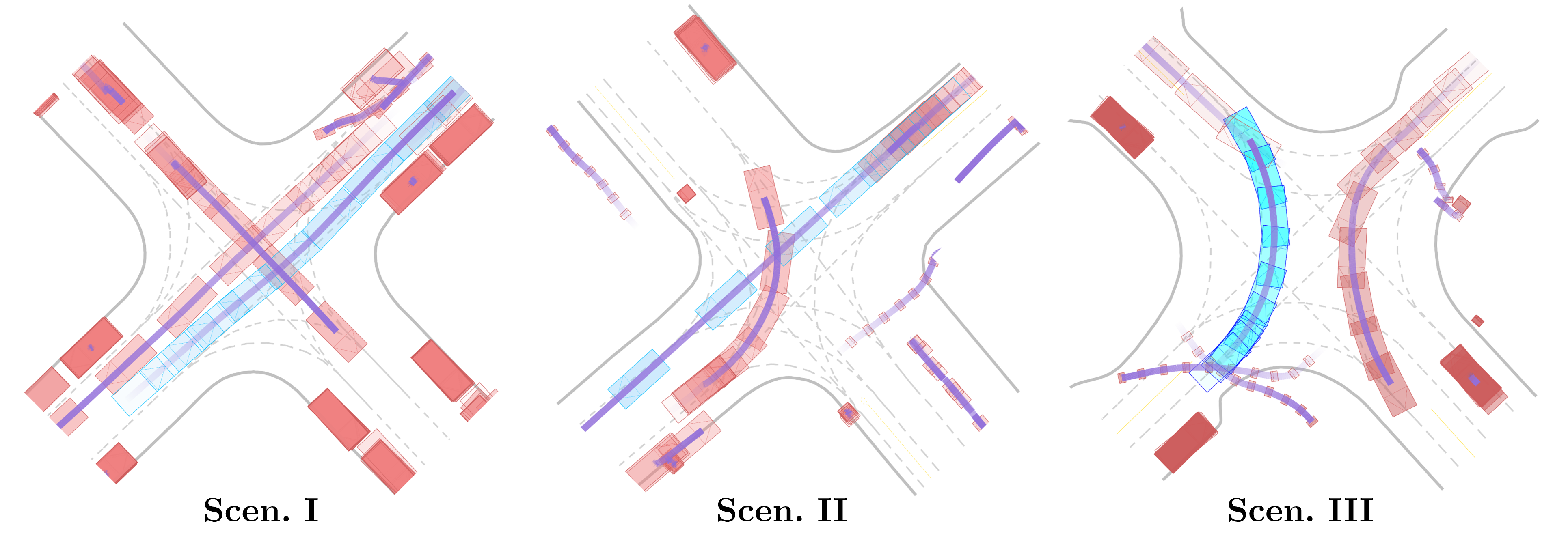}
     \caption{Intersection Scenarios selected from Argoverse 2 validation split for comparisons. The ego vehicle is colored in \textit{blue}. The trajectories are colored with fading \textit{purple}, and vehicles in key frames are visualized with fading colors, respectively. Scen. I: Three vehicles meet at the intersection. The vehicle from the top right exhibits a misleading intention of left turning, while the vehicle from the top left can easily be misidentified as taking the right-of-way. Scen. II: Two vehicles enter the intersection. The yielding intention of the bottom-left vehicle needs to be identified to determine the passing priority. Scen. III: Two vehicles approach the intersection. The vehicle's intention from the top right can easily be misidentified as going straight, which would influence the behavior of the ego vehicle during the unprotected left turn.}
     \label{fig:quant_illu}
     \vspace{-0.3cm}
\end{figure}

\begin{figure*}[h]
    \centering
    \includegraphics[width=0.95\textwidth]{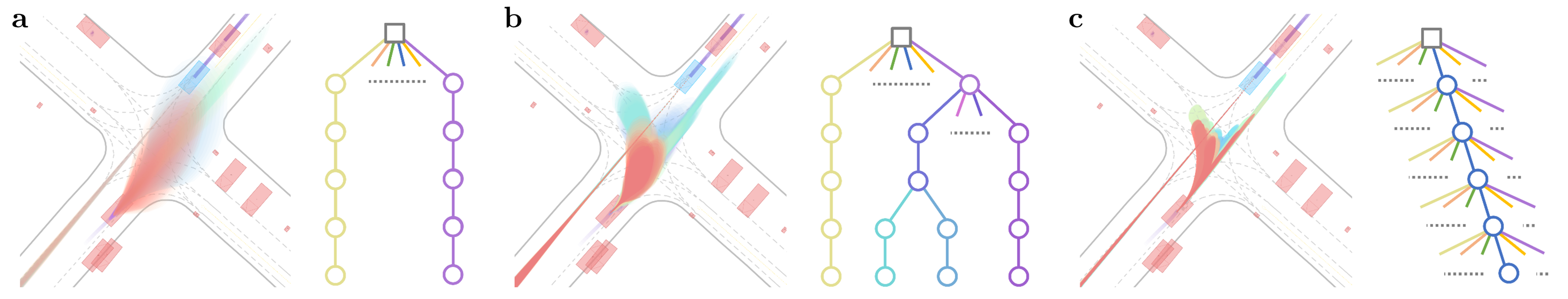}
    \caption{Snapshots of the effectiveness analysis in the intersection scenario. Predictions and decisions in the same scenarios are colored in the same color. The associated scenario trees are visualized on the right. (a) Single Shot: With no branching on the scenario tree, the predicted uncertainties of the oncoming agent increase sharply as entering the intersection. The predictions covering a large area of intersections place great challenges to the following contingency planning. (b) AIME: The prediction uncertainties are kept relatively low and the interaction patterns are clear within the sparse scenario tree. Compared with the Single Shot and Brute-force Search, AIME achieves good coverage on interaction modalities and better efficiency on scenario tree generations. (c) Brute-force Search: By branching with a fixed time gap, an exhaustive search is conducted to find future scenarios covering possible interaction modalities. The resulting scenario tree is complicated and computationally expensive to integrate with contingency planning.}
    \label{fig:aime_quan}
    \vspace{-0.3cm}
\end{figure*}

\subsubsection{Effectiveness analysis of AIME}
We conduct quantitative experiments to evaluate the effectiveness of AIME in three types of scenarios: highway, street and intersection. We randomly selected 50 data sequences of each type from the validation split of Argoverse 2. To obtain a complete set covering diverse interaction modalities for comparisons, a brute-force search (BF-SRCH) is adopted where predicted scenarios are expanded in a fixed-time-step manner. Meanwhile, a single shot (SS) method in which interaction modalities are obtained with one inference is provided as a baseline for comparison. For fair comparisons, neural networks and the hyper-parameters of interaction modality are consistent in the three methods. The averaged quantitative results are in~\autoref{table_2}. AIME demonstrates notable efficiency and modality coverage in the scenarios tested, outperforming the SS method which fails to cover enough interaction modalities due to the significant prediction uncertainties, and the BF-SRCH method, which suffers from inefficiency due to the expansion of superfluous scenarios. Visualizations of predicted scenarios of three methods in an intersection are shown in~\autoref{fig:aime_quan}.

\begin{table}[t]
\caption{Results of Effectiveness Study of AIME.}\label{table_2}
\centering
\setlength{\tabcolsep}{3.0mm}
\begin{tabular}{p{1.2cm}|p{1.2cm}|p{1.1cm}p{1.1cm}p{1.1cm}} 
\toprule
\multicolumn{2}{c}{Methods} &Cover.~($\%$)  &Scen.Num.        &Comp.Cost          \\ 
\midrule
\multirow{3}{*}{Highway}        & SS        & 93.6           & 6                & 1.0x          \\
                            & AIME      & 96.4           & 11.4             & 2.7x         \\ 
                            & BF-SRCH   & 100.0          & 7776             & 1974.8x       \\ 
\hline
\multirow{3}{*}{Street}        & SS        & 79.2           & 6                & 1.0x          \\
                            & AIME      & 88.9           & 34.3             & 4.6x          \\ 
                            & BF-SRCH   & 100.0          & 7776             & 2567.7x       \\ 

\hline
\multirow{3}{*}{Intersection}        & SS        & 17.8           & 6                & 1.0x          \\
                            & AIME      & 79.5           & 197.1            & 17.2x          \\ 
                            & BF-SRCH   & 100.0          & 7776             & 3179.5x       \\ 
\bottomrule
\end{tabular}
\vspace{-0.6cm}
\end{table}

\subsubsection{Quantitative comparisons in closed-loop simulations}
\begin{figure}
    \centering
    \includegraphics[width=0.48\textwidth]{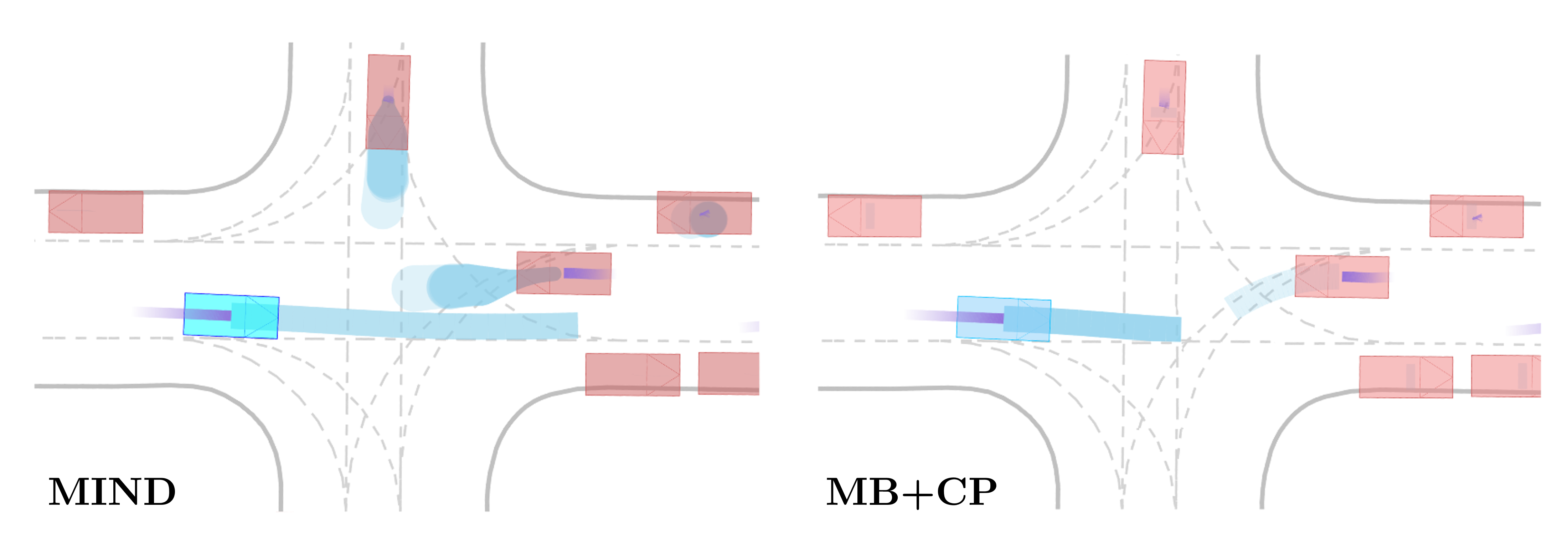}
     \caption{A snapshot of the closed-loop simulation in Scen.I. The historical trajectories are colored in the fading~\textit{purple}. The 2-sigma ellipses of the GMMs, predicted trajectories, and planned trajectories are visualized in~\textit{light blue}. MIND effectively identifies the interaction pattern where the oncoming vehicle may gradually advance to allow passing and executes a siding maneuver to overtake. Conversely, MB+CP predicts a less realistic scenario where both vehicles simultaneously attempt to give way, attributed to the handcrafted models' limited ability, leading to conservative slowing down.}
     \label{fig:snapshots}
     \vspace{-0.6cm}
\end{figure}

We further conduct closed-loop quantitative comparisons with the baselines across three typical scenarios selected from the aforementioned intersection group~\autoref{fig:quant_illu}. Since the selected scenarios are highly interactive, where the intentions and right-of-ways need to be determined promptly and precisely for safe and smooth navigation, the performances in these scenarios can showcase the superiority of MIND. As shown in \autoref{table_2}, MIND performs better in all three scenarios. Compared with MB+CP, MIND anticipates futures with better scene-consistent predictions and decisions, leading to more reasonable actions~\autoref{fig:snapshots}. Meanwhile, thanks to the multi-modal interactions explored with the guidance of AIME given the ``world dynamics", the uncertainties in each scenario are effectively narrowed down, enabling less conservative and more interaction-appropriate maneuvers of MIND.

\begin{table}[t]
\caption{Quantitative results of closed-loop simulations in three test driving scenarios. The better result is in \textbf{bold}.}\label{table_3}
\centering
\setlength{\tabcolsep}{3.0mm}
\begin{tabular}{l|l|ccc} 
\toprule
\multicolumn{2}{c}{Methods} &
\makecell{avgSpd\\$(m/s)\uparrow$} & 
\makecell{maxAbsAcc\\$(m/s^2)\downarrow$} &
\makecell{rmsAcc\\$(m/s^2)\downarrow$} \\ 
\midrule
\multirow{3}{*}{Scen. I}    & MB+CP         & 4.00          & 1.53              & 0.75 \\
                            & NN+CP         & 3.42          & 1.65              & 0.84 \\
                            & MIND          & \textbf{4.25} & \textbf{0.89}    & \textbf{0.59} \\ 
\hline
\multirow{3}{*}{Scen. II}   & MB+CP         & 3.83          & 1.01              & 0.82 \\ 
                            & NN+CP         & 3.25          & 1.30              & 0.98 \\
                            & MIND          & \textbf{4.14} & \textbf{0.99}   & \textbf{0.76}\\
\hline
\multirow{3}{*}{Scen. III}  & MB+CP         & 2.24          & 1.22              & 0.67 \\ 
                            & NN+CP         & 2.42          & 1.43              & 0.74 \\
                            & MIND          & \textbf{2.63} & \textbf{1.04}     & \textbf{0.66} \\
\bottomrule
\end{tabular}
\vspace{-0.4cm}
\end{table}

\subsubsection{Qualitative results of closed-loop simulations}
\begin{figure*}[t]
    \centering
    \includegraphics[width=\textwidth]{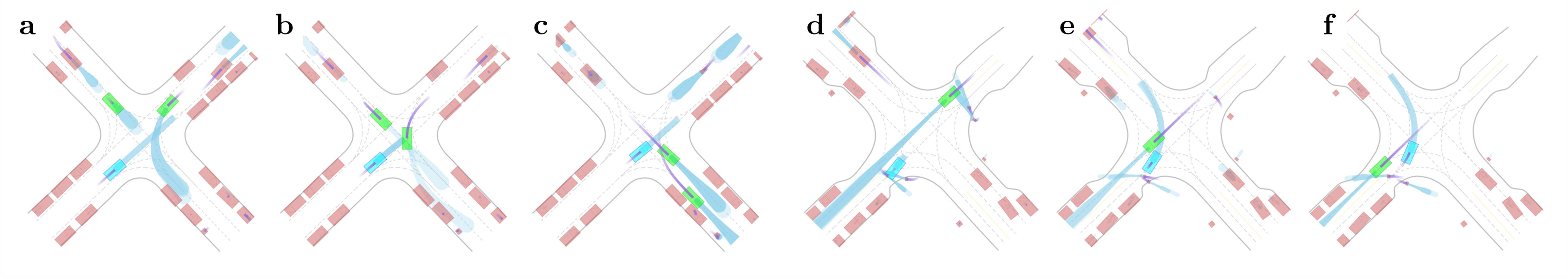}
    \caption{Snapshots of the closed-loop simulations with adversarial agents. The adversarial agents are colored in \textit{green}. (a-c) Scen.I with two adversarial agents: The agent from the top right aggressively swerves to the left with a sudden acceleration. Meanwhile, the agent on the top left accelerates to get the right-of-way. (a) The MIND first anticipates the top-right agent's aggressive left-turning intention and the agent's yielding intention on the left. It slowly moves forward after yielding to the left-turning maneuver. (b) MIND notices the sudden acceleration of the top-left vehicle and quickly makes the yielding decision. (c) MIND accelerates to leave after yielding. (d-e) Scen.III with one smart agent: The agent from the top right changes its intention to go straight with aggressive acceleration instead of turning left. (d) MIND performs a human-like creeping behavior while predicting the agent's determining will to go straight. (e-f) MIND resumes the left-turning decision in advance when it predicts that the agent is leaving the intersection.}
    \label{fig:qual_exp_1}
    \vspace{-0.5cm}
\end{figure*}

We conduct qualitative experiments to evaluate MIND's capability of interaction handling in multi-agent scenarios with diverse behaviors. To achieve this, we modify the scenarios from the quantitative analysis by incorporating adversarial agents that exhibit dynamic actions. Additionally, taking the idea of worst-case analysis, we heighten the risk and urgency of the scenarios by assigning aggressive policies (such as sudden accelerations, abrupt changes in direction from straight to turning or forcing right-of-way changes) to the adversarial agents. The results, as illustrated in~\autoref{fig:qual_exp_1}, reveal MIND's capacity to adapt to the evolving intentions of other agents and make considerate decisions with human-like behaviors, demonstrating its adaptability.

\subsubsection{Qualitative results of conditioned scenario predictions}
\begin{figure}[t]
    \centering
    \includegraphics[width=0.485\textwidth]{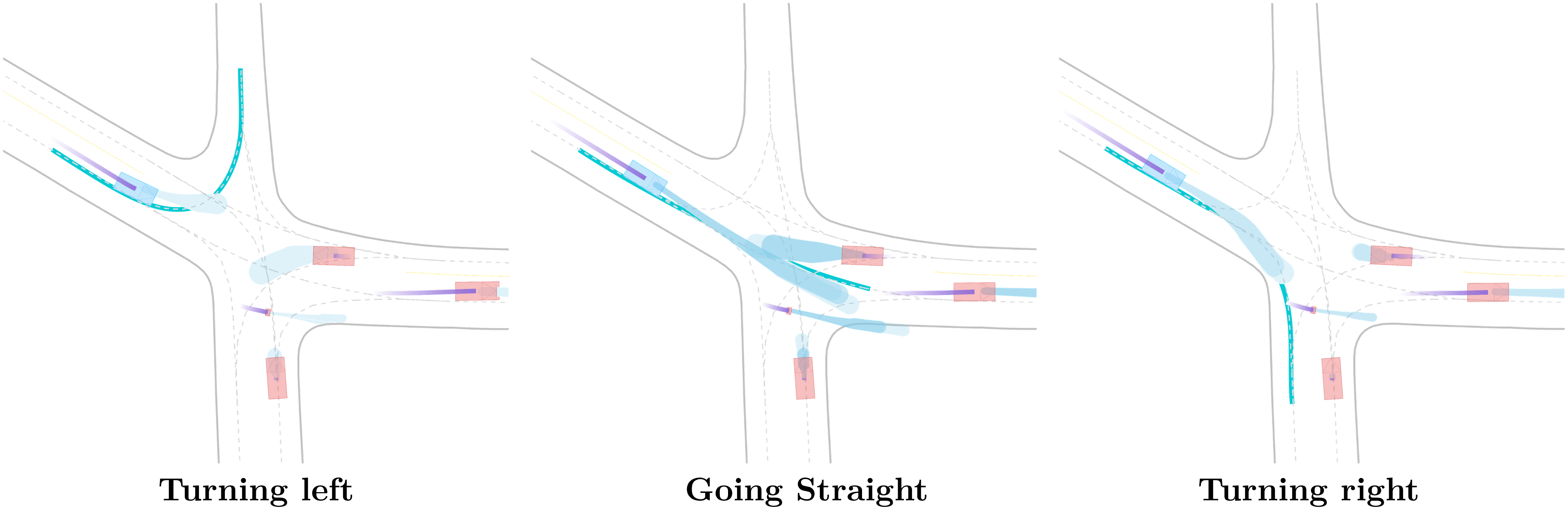}
     \caption{Qualitative results of scenario predictions conditioned on different high-level commands. In the ``Turning left" prediction, The predicted ego decision exhibits a deceleration maneuver due to the current high speed while the prediction of the oncoming vehicle continues the left turn without yielding to the ego vehicle. In the ``Going straight" prediction, The oncoming vehicle is predicted to move forward, waiting for the ego vehicle to pass the intersection. In the ``Turning right" prediction, The ego decision first turns to the right and then yields at the crossing pedestrian. The oncoming vehicle is predicted to wait and yield at the ego vehicle.}
     \label{fig:qual_exp_2}
     \vspace{-0.5cm}
\end{figure}

To illustrate the network's proficiency in generating distinct predictions tailored to different planning objectives, we conduct a qualitative analysis with different high-level commands as conditioned inputs. The inputted routes vary based on the specified high-level commands throughout these tests, yet the historical data and map information remain unchanged. As depicted in~\autoref{fig:qual_exp_2}, the proposed network successfully produces a variety of plausible scenarios tailored to each command, highlighting its capability for conditioned prediction and its flexibility in responding to various commands.

\section{Conclusion and Future Work}\label{sec:concl}
We introduce the MIND framework as a comprehensive approach for simultaneous prediction and decision-making in autonomous driving within dynamic interactive settings. The framework systematically combines a scenario prediction network, adaptive interaction modality exploration mechanism, and contingency planning to generate reasonable behaviors in complex interaction scenarios while handling multi-modal future evolutions. Extensive quantitative comparisons against state-of-the-art and qualitative experiments have demonstrated the superiority of our approach. Moving forward, we aim to extend our proposed framework to real-world applications.


\bibliographystyle{IEEEtran}
\bibliography{ref}

\begin{thebibliography}{10}
\providecommand{\url}[1]{#1}
\csname url@samestyle\endcsname
\providecommand{\newblock}{\relax}
\providecommand{\bibinfo}[2]{#2}
\providecommand{\BIBentrySTDinterwordspacing}{\spaceskip=0pt\relax}
\providecommand{\BIBentryALTinterwordstretchfactor}{4}
\providecommand{\BIBentryALTinterwordspacing}{\spaceskip=\fontdimen2\font plus
\BIBentryALTinterwordstretchfactor\fontdimen3\font minus \fontdimen4\font\relax}
\providecommand{\BIBforeignlanguage}[2]{{%
\expandafter\ifx\csname l@#1\endcsname\relax
\typeout{** WARNING: IEEEtran.bst: No hyphenation pattern has been}%
\typeout{** loaded for the language `#1'. Using the pattern for}%
\typeout{** the default language instead.}%
\else
\language=\csname l@#1\endcsname
\fi
#2}}
\providecommand{\BIBdecl}{\relax}
\BIBdecl

\bibitem{ding2021epsilon}
W.~Ding \emph{et~al.}, ``Epsilon: An efficient planning system for automated vehicles in highly interactive environments,'' \emph{{IEEE} Trans. on Robot.}, vol.~38, no.~2, pp. 1118--1138, 2021.

\bibitem{cui2019multimodal}
H.~Cui \emph{et~al.}, ``Multimodal trajectory predictions for autonomous driving using deep convolutional networks,'' in \emph{ICRA. IEEE}, 2019.

\bibitem{hubmann2018automated}
C.~Hubmann \emph{et~al.}, ``Automated driving in uncertain environments: Planning with interaction and uncertain maneuver prediction,'' \emph{{IEEE} Trans. on Intel. Veh.}, vol.~3, no.~1, pp. 5--17, 2018.

\bibitem{zeng2020dsdnet}
W.~Zeng \emph{et~al.}, ``{DSDNet}: Deep structured self-driving network,'' in \emph{ECCV. Springer}.\hskip 1em plus 0.5em minus 0.4em\relax Springer, 2020, pp. 156--172.

\bibitem{cui2021lookout}
A.~Cui \emph{et~al.}, ``Lookout: Diverse multi-future prediction and planning for self-driving,'' in \emph{ICCV. IEEE}, 2021, pp. 16\,107--16\,116.

\bibitem{casas2021mp3}
S.~Casas \emph{et~al.}, ``{MP3}: A unified model to map, perceive, predict and plan,'' in \emph{CVPR}, 2021, pp. 14\,403--14\,412.

\bibitem{rhinehart2019precog}
N.~Rhinehart \emph{et~al.}, ``{PRECOG}: Prediction conditioned on goals in visual multi-agent settings,'' in \emph{ICCV. IEEE}, 2019, pp. 2821--2830.

\bibitem{song2020pip}
H.~Song \emph{et~al.}, ``{PiP}: Planning-informed trajectory prediction for autonomous driving,'' in \emph{ECCV. Springer}.\hskip 1em plus 0.5em minus 0.4em\relax Springer, 2020, pp. 598--614.

\bibitem{salzmann2020trajectron++}
T.~Salzmann \emph{et~al.}, ``Trajectron++: Dynamically-feasible trajectory forecasting with heterogeneous data,'' in \emph{ECCV. Springer}, 2020, pp. 683--700.

\bibitem{casas2020implicit}
S.~Casas \emph{et~al.}, ``Implicit latent variable model for scene-consistent motion forecasting,'' in \emph{ECCV. Springer}.\hskip 1em plus 0.5em minus 0.4em\relax Springer, 2020, pp. 624--641.

\bibitem{girgis2021latent}
R.~Girgis \emph{et~al.}, ``Latent variable sequential set transformers for joint multi-agent motion prediction,'' \emph{arXiv preprint arXiv:2104.00563}, 2021.

\bibitem{ngiam2021scene}
J.~Ngiam \emph{et~al.}, ``{Scene} {Transformer}: A unified multi-task model for behavior prediction and planning,'' \emph{arXiv preprint arXiv:2106.08417}, vol.~2, no.~7, 2021.

\bibitem{kang2023ffinet}
M.~Kang \emph{et~al.}, ``{FFINet}: Future feedback interaction network for motion forecasting,'' \emph{arXiv preprint arXiv:2311.04512}, 2023.

\bibitem{hagedorn2023rethinking}
S.~Hagedorn \emph{et~al.}, ``Rethinking integration of prediction and planning in deep learning-based automated driving systems: a review,'' \emph{arXiv preprint arXiv:2308.05731}, 2023.

\bibitem{chen2023tree}
Y.~Chen \emph{et~al.}, ``Tree-structured policy planning with learned behavior models,'' in \emph{ICRA. IEEE}.\hskip 1em plus 0.5em minus 0.4em\relax IEEE, 2023, pp. 7902--7908.

\bibitem{luo2018porca}
Y.~Luo \emph{et~al.}, ``{PORCA}: Modeling and planning for autonomous driving among many pedestrians,'' \emph{{IEEE} Robot. Autom. Lett.}, 2018.

\bibitem{huang2023dtpp}
Z.~Huang \emph{et~al.}, ``Dtpp: Differentiable joint conditional prediction and cost evaluation for tree policy planning in autonomous driving,'' in \emph{ICRA. IEEE}, 2024.

\bibitem{chekroun2023mbappe}
R.~Chekroun \emph{et~al.}, ``{MBAPPE}: {MCTS}-built-around prediction for planning explicitly,'' \emph{arXiv preprint arXiv:2309.08452}, 2023.

\bibitem{li2023marc}
T.~Li \emph{et~al.}, ``{MARC}: Multipolicy and risk-aware contingency planning for autonomous driving,'' \emph{{IEEE} Robot. Autom. Lett.}, 2023.

\bibitem{liang2020learning}
M.~Liang \emph{et~al.}, ``Learning lane graph representations for motion forecasting,'' in \emph{ECCV. Springer}.\hskip 1em plus 0.5em minus 0.4em\relax Springer, 2020, pp. 541--556.

\bibitem{gao2020vectornet}
J.~Gao \emph{et~al.}, ``{VectorNet}: Encoding hd maps and agent dynamics from vectorized representation,'' in \emph{CVPR}, 2020, pp. 11\,525--11\,533.

\bibitem{varadarajan2022multipath++}
B.~Varadarajan \emph{et~al.}, ``{MultiPath}++: Efficient information fusion and trajectory aggregation for behavior prediction,'' in \emph{ICRA. IEEE}, 2022.

\bibitem{luo2023jfp}
W.~Luo \emph{et~al.}, ``Jfp: Joint future prediction with interactive multi-agent modeling for autonomous driving,'' in \emph{CoRL}, 2023, pp. 1457--1467.

\bibitem{sun2022m2i}
Q.~Sun \emph{et~al.}, ``{M2I}: From factored marginal trajectory prediction to interactive prediction,'' in \emph{CVPR}, 2022, pp. 6543--6552.

\bibitem{rowe2023fjmp}
L.~Rowe \emph{et~al.}, ``{FJMP}: Factorized joint multi-agent motion prediction over learned directed acyclic interaction graphs,'' in \emph{CVPR}, 2023.

\bibitem{zhang2024simpl}
L.~Zhang \emph{et~al.}, ``Simpl: A simple and efficient multi-agent motion prediction baseline for autonomous driving,'' \emph{{IEEE} Robot. Autom. Lett.}, 2024.

\bibitem{cunningham2015mpdm}
A.~G. Cunningham \emph{et~al.}, ``{MPDM}: Multipolicy decision-making in dynamic, uncertain environments for autonomous driving,'' in \emph{ICRA. IEEE}, 2015, pp. 1670--1677.

\bibitem{zhang2020efficient}
L.~Zhang \emph{et~al.}, ``Efficient uncertainty-aware decision-making for automated driving using guided branching,'' in \emph{ICRA. IEEE}, 2020, pp. 3291--3297.

\bibitem{hardy2013contingency}
J.~Hardy and M.~Campbell, ``Contingency planning over probabilistic obstacle predictions for autonomous road vehicles,'' \emph{{IEEE} Trans. on Robot.}, vol.~29, no.~4, pp. 913--929, 2013.

\bibitem{chen2023interactive}
Y.~Chen \emph{et~al.}, ``Interactive joint planning for autonomous vehicles,'' \emph{{IEEE} Robot. Autom. Lett.}, 2023.

\bibitem{wang2023interaction}
R.~Wang \emph{et~al.}, ``Interaction-aware model predictive control for autonomous driving,'' in \emph{ECC. IEEE}, pp. 1--6.

\bibitem{mayne1966second}
D.~Mayne, ``A second-order gradient method for determining optimal trajectories of non-linear discrete-time systems,'' \emph{Intl. J. Ctrl.}, vol.~3, no.~1, pp. 85--95, 1966.

\bibitem{wilson2023argoverse}
B.~Wilson \emph{et~al.}, ``Argoverse 2: Next generation datasets for self-driving perception and forecasting,'' \emph{arXiv preprint arXiv:2301.00493}, 2023.

\end{thebibliography}

\end{document}